\documentclass{article}

\usepackage[final]{nips_2017}

\usepackage[utf8]{inputenc} 
\usepackage[T1]{fontenc}    
\usepackage{hyperref}       
\usepackage{url}            
\usepackage{booktabs}       
\usepackage{amsfonts}       
\usepackage{amsmath}
\usepackage{graphicx}
\usepackage{subcaption}
\usepackage{multirow}
\usepackage{nicefrac}       
\usepackage{microtype}      
\usepackage{color}
\usepackage{multirow}

\newcommand{\mb}{\mathbf}

\title{CMU LiveMedQA at TREC 2017 LiveQA: A Consumer Health Question Answering System}

\author{
  Yuan Yang \\
  \And
  Jingcheng Yu \\
  \And
  Ye Hu \\
  \And
  Xiaoyao Xu \\
  \And
  Eric Nyberg \\
  \AND
  \vspace{-.5cm}
  \\
  Language Technologies Institute \\
  Carnegie Mellon University \\
  Pittsburgh, PA 15213, USA \\
  \texttt{\{yuany2, jingchey, yhu1, xiaoyaox\}@andrew.cmu.edu} \\
  \texttt{ehn@cs.cmu.edu} \\
}

\begin{document}

\maketitle

\begin{abstract}
In this paper, we present LiveMedQA, a question answering system that is optimized for consumer health question. On top of the general QA system pipeline, we introduce several new features that aim to exploit domain-specific knowledge and entity structures for better performance. This includes a question type/focus analyzer based on deep text classification model, a tree-based knowledge graph for answer generation and a complementary structure-aware searcher for answer retrieval. LiveMedQA system is evaluated in the TREC 2017 LiveQA medical subtask, where it received an average score of 0.356 on a 3 point scale. Evaluation results revealed 3 substantial drawbacks in current LiveMedQA system, based on which we provide a detailed discussion and propose a few solutions that constitute the main focus of our subsequent work.
\end{abstract}

\section{Introduction}

The TREC LiveQA Challenge~\cite{agichtein2015overview} has been conducted since 2015, and focuses on generating answers for human questions in real time. Starting in 2017, a LiveQA medical subtask was added, which focuses on forming answers for questions involving consumer-generated inquiries about personal health status, treatment recommendations and general information/consultation on other common medical entities, i.e. \textit{consumer health questions}. The subtask shares a common format and requirements with the main track, where upon receiving a question request, each participant system is required to generate an answer of up to 1000 characters in 60 seconds.

In parallel with the continued efforts of the CMU Open Advancement of Question Answering (OAQA) group in the general LiveQA task~\cite{wang2015cmu,wang2016cmu}, the CMU LiveMedQA team aims to develop a novel QA system specialized in tackling the health subtask challenge. This system, while sharing a similar input/output pipeline with the general approach, differs from domain-general systems in several aspects: (1) the LiveMedQA system builds and maintains a medical domain knowledge graph that organizes information for medical entities as trees, with the entity as root and its attributes as leaves; (2) LiveMedQA infers question type and focus with a deep learning model, and utilizes them during processing with a set of question types that is slightly different from those in the official development set; each question type can be directly mapped to certain attributes in the knowledge tree; and (3) the system supports merging and re-ranking candidates retrieved from multiple sources, including web search and a built-in knowledge graph.

During the official evaluation, the LiveMedQA system received 104 questions and successfully submitted answers for 103 of them. System responses were judged by human assessors on a 4-level Likert scale. The LiveMedQA system achieved an average score of 0.356, which is comparatively lower than the median of 0.552. In the following section, we first present the detailed design of the LiveMedQA system; then in section 3, we present evaluation results and analyze possible sources of error in the system.

\section{System Architecture}

\begin{figure}[tb]
\centering
\includegraphics[width=\columnwidth, trim={0cm 4.5cm 2cm 3cm}, clip]{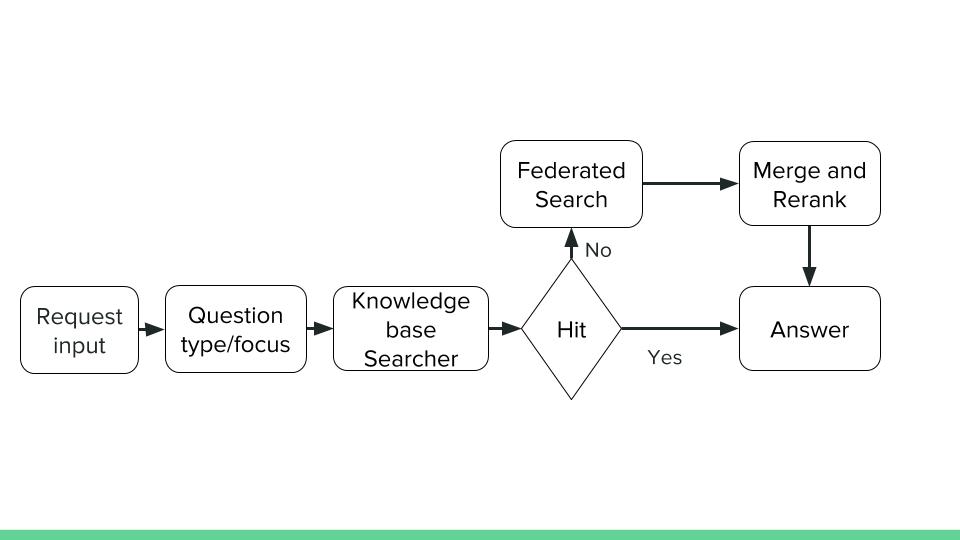}
\caption{Overview of LiveMedQA system architecture.}
\label{fig:pipeline}
\end{figure}

Unlike general questions, consumer health questions usually can be seen as a query with respect to a set of medical entities and their corresponding attributes. For example, the question ``How do I know if I have diabetes?'' can be formed as a query for retrieving the ``Symptom'' attribute associated with entity ``Diabetes''. For simplicity of subsequent implementation, we assume each question contains exactly one such entity with one or more attributes associated with it. Apparently, this assumption does not hold in many situations, such as questions about drug interactions, which usually involves comparison between two entities. But after inspecting the training set, we found this assumption cover most of the cases. And we will discuss its effect on the test dataset in the result section.

Thus the original task can be decomposed into several stages: first, recognizing medical entity and attributes; second, retrieving contents of the associated contents, which involves building a knowledge graph that hold these information; and third, ranking contents and generating final answer.

The LiveMedQA system consists of several modules each serves to process the question at the corresponding stage. As shown in Fig.\ref{fig:pipeline}, upon receiving a request, system will analyze its question type and focus which are taken as the entity and attributes: the former is achieved by utilizing a text classification convolutional neural network (CNN) model trained on the official development set and a public dataset~\cite{roberts2014automatically}, whereas the latter is done by resorting to traditional NLP toolkit such as NLTK~\cite{bird2006nltk}. 

In LiveMedQA, these two features play an important role in the remaining components, specifically in its knowledge graph module. This module maintains an array of medical entities ranging from specific diseases such as diabetes to clinical procedures and medications. Each entity contains a set of attributes that characterizing the object in different aspects. Attributes are organized in a tree structure that preserves the medical domain-specific concept hierarchy: entity name is put in the root, and each leave for each attribute. Attributes belonging to the same subtree share the same high-level concept. For example, a subtree with root ``Treatment'' stores leave nodes such as ``Nonpharmacologic Therapy'' and  ``Medication''. And in order to search contents in the knowledge graph, we implement a structure-aware searcher that makes use of regular expression and a \textit{Lucene} BM25 search engine for matching the medical entity; as well as an attribute lookup table for matching the type to certain entity attribute.

In parallel with the offline knowledge base, LiveMedQA also retrieves candidates from several online resources such as Yahoo!Answer\footnote{https://answers.yahoo.com/}, and MedlinePlus Web Service\footnote{https://medlineplus.gov/webservices.html}. The system turns to retrieve from these sources whenever it cannot find good candidates in knowledge base.

The following sections provide detailed discussion on each of these modules.

\subsection{Question focus extraction}

Extracting question focus (i.e. medical entity) is more straightforward than that in general domain, since medical entities usually have a significantly higher TF-IDF scores than the rest of the context. To do this, we extract all nouns and noun phrases using NLTK toolkit. Then each candidate ranked by the average TF-IDF scores of all its containing tokens (nouns contains exactly one token). Token TF-IDF scores are computed from a background corpus built upon MedlinePlus Health Topics\footnote{https://medlineplus.gov/xml.html}.

\subsection{Question type inference with text classification CNN}

The official training dataset contains 23 different question types. While some of them such as ``Cause'', ``Complication'' and ``Treatment'' can be easily linked to an entity, types such as ``Association'' and ``Organization'' are usually ambiguous and can potentially contain overlapping information. Thus we condense the type set into 10 categories: Treatment, Information, Susceptibility, Prognosis, Symptom, Diagnosis, Cause, Organization, Drug Information, and Drug Interaction. Another consideration is that original types are highly skewed in frequency, which usually leads machine learning model ignoring the rare classes.

\begin{figure}[tb]
\centering
\includegraphics[width=.8\columnwidth, trim={0cm 0cm 0cm 0cm}, clip]{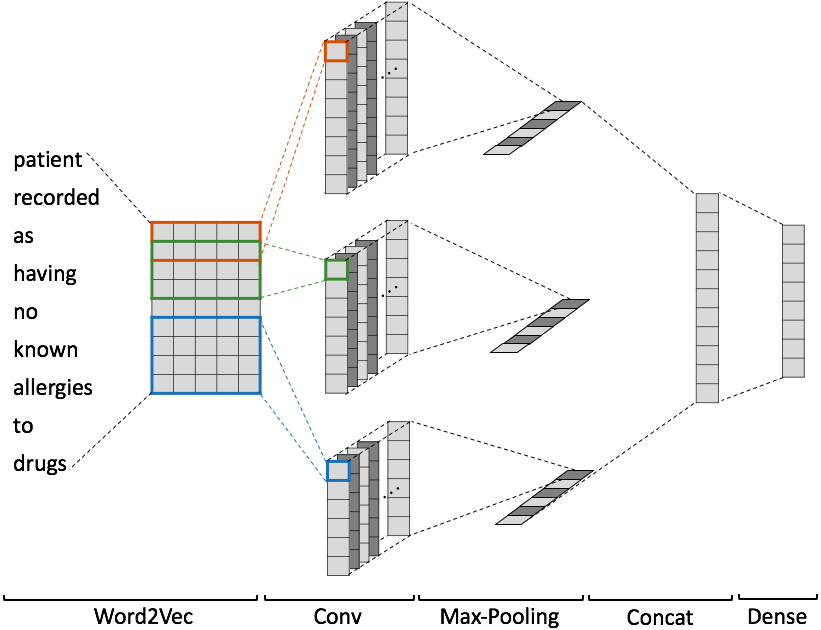}
\caption{The overall architecture of CNN model with 3 stacks of convolution filters, each with window sizes of 2,3 and 4. Diagram adopted from Kim et al.~\cite{kim2014convolutional}.
}
\label{fig:arch}
\end{figure}

Predicting the question type can formalized as a multi-label text classification problem. And here we utilize a CNN model~\cite{kim2014convolutional} to achieve this. Its architecture is shown in Fig.\ref{fig:arch}. Formally, each input question is represented as a sequence of words $t_1, t_2, \cdots, t_l$, where $t_i$ denotes the $i$-th word with length of $l$. Words are transformed using \textit{Word2Vec}~\cite{mikolov2013distributed}. The idea here is to learn a distributed vector representation for each word. Specifically, it learns a parameter matrix $\mb{T}\in\mathbb{R}^{v\times h}$, where $v$ is the size of vocabulary and $h$ denotes the dimension of embedding vectors, so that $i$-th row vector corresponds to the embedding of word $i$. Eventually, the question is transformed to a $l$-by-$h$ matrix $\mb{D}$. 

This matrix is then fed into the convolution layer that learns a set of filters to capture the local semantic information in the question. For a convolution filter with window size of $n$, we parameterize it into a $n$-by-$h$ weight matrix $\mb{W}$ and a bias term $b$. Note that the width of the filter is always set to $h$, since each word is represented by a complete $h$-dimensional vector. Convolution is performed on each $n$-by-$h$ chunk $\mb{D}_{i:i+n-1}$ in matrix $\mb{D}$. The output is a new feature value computed as 
$$c_i=f(\langle \mb{W}, \mb{D}_{i:i+n-1}\rangle +b),$$ 
where $\langle\cdot, \cdot, \rangle$ denotes matrix inner product and $f(\cdot)$ denotes a nonlinear activation function such as rectified linear. As shown in Fig.\ref{fig:arch}, the input matrix is processed by a few stacks of filters, where filters in different stack have different window size, and each filter possesses a set of independent parameters. The idea is to capture a variety of features of variable length. The following pooling layer selects the one with largest activation value from each filter. The pooled values of all filters are then concatenated together into a complete representation $\mb{x}$.

Finally, a fully connected layer with linear output is used for predicting the question type. Each hidden unit $i$ is parameterized as a weight vector $\mb{u}_i$ and a bias term $d_i$. Given representation $\mb{x}$, hidden unit $i$ produces an activation:
$$z_i=f(\mb{u}_i^\top\mb{x}+d_i).$$ 
All latent units are then mixed together with another weight vector $\mb{o}_i$ and a bias term $e_i$, where we can represent the probability of question type $i$ as
$$p_i=\sigma(\mb{o}_i^\top\mb{z}+e_i),$$
where $\sigma(\cdot)$ is the sigmoid function.

\textbf{Training}: We set the loss function to the cross entropy between prediction $\mb{p}^{(j)}$ and the ground-truth $\mb{y}^{(j)}$, which is given as
$$-\sum_{j=1}^{m}\sum^k_{i=1} y_{i}^{(j)} \log p_i^{(j)} + (1-y_{i}^{(j)})\log(1 - p_i^{(j)}).$$ 
We then apply \textit{stochastic gradient descent} method to update all learnable parameters in the network in order to minimize this loss. 
We use ReLu function for all non-linear transformation, and perform batch normalization~\cite{ioffe2015batch} at the end of each layer. Dropout~\cite{srivastava2014dropout} is used on representation vector $\mb{x}$ and L2 regularization is applied on all learnable parameters.

\subsection{Building Knowledge Graph and Structure-Aware Searcher}

\begin{figure}[tb]
\centering
\includegraphics[width=\columnwidth, trim={0cm 2cm 2cm 2cm}, clip]{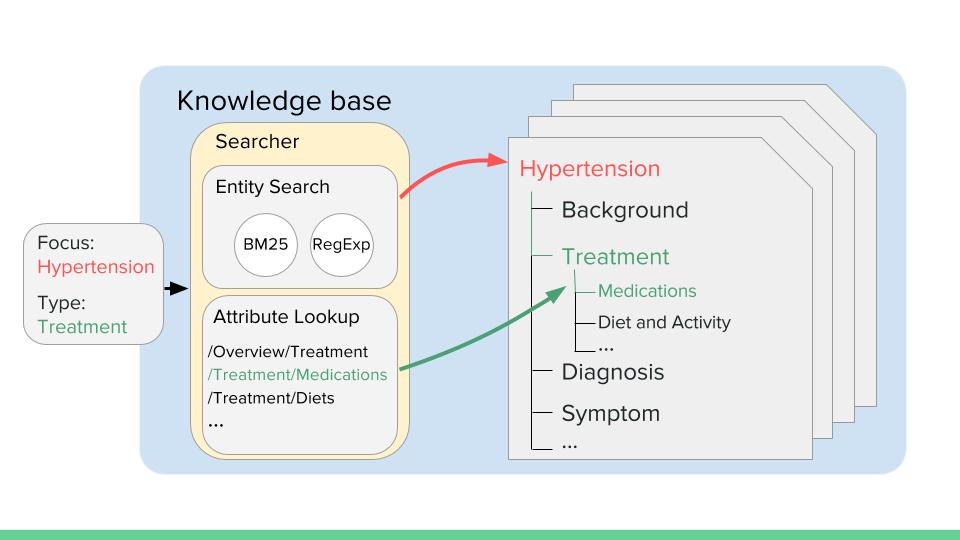}
\caption{Overview of system knowledge graph and structure-aware searcher design. It illustrates the process of finding the ``Treatment'' attribute for entity ``Hypertension''.}
\label{fig:knowledge_graph}
\end{figure}

Retrieving contents using focus and types in LiveMedQA requires the system to store relevant knowledge with structured format and a dedicated searcher to match specific contents. These two components are shown in Fig.\ref{fig:knowledge_graph}. We organize the knowledge base as a collection of trees. Each tree represents a specific medical entity such as ``Hypertension''. Attributes associated with an entity can be then organized as a hierarchy of concepts, where nodes close to the root are more abstract than those on the leaves. Using a deep hierarchy is more beneficial than a flattened structure, since it allows us to match relevant attributes in the same subtree when primary attribute does not exist.

To build such graph, we first consult physicians and gather a list of common medical entities such as diseases, medications and clinical procedures, and define several hierarchy templates for different types of entities (as shown in Fig.\ref{fig:knowledge_graph}). We then implemented a web crawler that crawls relevant online webpages. We assume these webpages are organized as sequences of headings and corresponding contents and implemented text processing scripts for screening and matching each heading into the attribute slots.

Searching in the graph involves matching the correct entity and retrieving correct attribute contents from its tree. For the first part, we implemented a searcher that matches entity name with regular expression, as well as a \textit{Lucene} BM25 search engine that matches the flattened contents within the tree. Once the tree is located, the searcher tries to traverse the tree with a set of pre-defined paths specified by an attribute lookup table. This table essentially encapsulates the difference between templates and allows us to encode preference over some certain attributes.

\section{Results}

\begin{table}[tb]
\centering
\begin{tabular}{|l|c|lll|lll|}
\hline
\multirow{2}{*}{System} & \multirow{2}{*}{Avg Score} & \multicolumn{3}{c|}{Success@}                                          & \multicolumn{3}{c|}{Precision@}                                        \\
                        &                            & \multicolumn{1}{c}{2} & \multicolumn{1}{c}{3} & \multicolumn{1}{c|}{4} & \multicolumn{1}{c}{2} & \multicolumn{1}{c}{3} & \multicolumn{1}{c|}{4} \\ \hline
CMU-LiveMedQA           & 0.356                      & 0.216                 & 0.137                 & 0.000                  & 0.218                 & 0.139                 & 0.000                  \\
Median                  & 0.552                     & 0.245                 & 0.142                 & 0.059                  & 0.331                 & 0.178                 & 0.078                  \\
Max                     & 0.794                      & 0.392                 & 0.265                 & 0.098                  & 0.429                 & 0.273                 & 0.111                  \\ \hline
\end{tabular}
\vspace{.3cm}
\caption{Evaluation results on LiveMedQA system together with median and max performance.}
\label{tab:res}
\end{table}

During official evaluation, in total 104 questions are received by the system and results of 103 of them are successfully submitted to the track server. Individual answers are evaluated by a 4-level Likert scale: excellent (4), good (3), fair (2) and bad (1). System overall performance is measured by 3 approaches: 1) the average score of all answers generated; 2) Success@$i$, which is the percentage of answers being scored as $i$ among all questions received; and 3) Precision@$i$, which is the percentage of answers being scored as $i$ among all answers generated. Evaluation results are shown in Tab.\ref{tab:res}. We find LiveMedQA system generally performs worse than average, and receives scores below median. 


\section{Discussion}

We conducted a series of error analysis and concluded 3 main factors that possibly result in the low performance. First, a number of irrelevant results are caused by the miss-classification of question type. Since question type and focus play the central role in subsequent process, LiveMedQA system is very sensitive to the error generated at this stage, i.e. the performance of CNN model. This drawback becomes severe in this task, since CNN model is trained only on a tiny dataset with highly biased class-frequency (even after augmentation), which potentially leads to issues such as over-fitting and random behavior on rare classes. 

Another category of errors is introduced in the knowledge base module. Specifically, some errors are caused by the inaccurate matches of contents either using hard match or BM25 searcher; others are simply because the target entity does not exist and a similar entity is returned. Although LiveMedQA alleviates this type of error by introducing federated search with 2 other online sources, it is set up to highly prefer the candidates from knowledge base over the online sources (we found this leads to a better performance on training dataset). Thus, the wrong candidate is likely to be returned as the final results.

Finally, we observed a significant distribution shift of target labels in test dataset, where majority of them are medication related questions, including interaction, side effects and contraindications. Due to the assumption that only one entity exists per question, LiveMedQA system is almost certain to fail in this type of question. On the other hand, we do observe more a consistent good performance over disease related questions, since they generally follow the assumption.

\subsection{System remarks and future work}

The LiveMedQA system serves as the very first attempt by OAQA group in developing a QA systems specialized for consumer health questions. Although, during this year's evaluation, it turns out to be less competitive than vanilla QA strategies, we do observe a strong potentials of this design paradigm over the general purpose QA system framework. Here, we would like to note several chances where we believe the system can be substantially improved.

As the current system performance is heavily bounded by the type/focus module and the ad-hoc single-entity assumption, the first two chances to improve are: (1) training a more robust type prediction model and (2) relaxing the assumption to multiple entities. Achieving (1) ideally just requires more training data, but since they are generally difficult to acquire and label, one can resort to methods such transfer learning and domain adaptation in order to take advantage of the data outside the target domain. On the other hand, relaxing the assumption essentially implies that some of the entity attributes (e.g. drug interaction) are no longer associated with one by multiple entities, namely \textit{relations}. In that case, these relations can be explicitly modeled using probabilistic models such as Conditional Random Field (CRF), and correspondingly, the knowledge graph will become trees with a few leaves connected.

On the other hand, the knowledge graph can be also significantly improved, as contents are currently extracted from online sources and added to the graph without validation; and stored in only handful templates. One can implement more fine-grained crawlers for specific websites and limit search space to those trust sources.



\section{Conclusion}

In this paper, we present a question answering system that is optimized for consumer health question, namely LiveMedQA. On top of the general QA system pipeline, LiveMedQA is equipped with several novel features including a question type/focus analyzer, a domain-specific knowledge base and a structure-aware searcher. Evaluation results revealed 3 substantial drawbacks in current LiveMedQA system, based on which we provide a detailed discussion and propose a few viable solutions which constitute the main focus of our subsequent work.

\section*{References}

\small

\renewcommand{\bibsection}{}
\nocite{*}
\bibliography{bibliography}

\begin{thebibliography}{9}
\providecommand{\natexlab}[1]{#1}
\providecommand{\url}[1]{\texttt{#1}}
\expandafter\ifx\csname urlstyle\endcsname\relax
  \providecommand{\doi}[1]{doi: #1}\else
  \providecommand{\doi}{doi: \begingroup \urlstyle{rm}\Url}\fi

\bibitem[Agichtein et~al.(2015)Agichtein, Carmel, Pelleg, Pinter, and
  Harman]{agichtein2015overview}
Eugene Agichtein, David Carmel, Dan Pelleg, Yuval Pinter, and Donna Harman.
\newblock Overview of the trec 2015 liveqa track.
\newblock In \emph{TREC}, 2015.

\bibitem[Bird(2006)]{bird2006nltk}
Steven Bird.
\newblock Nltk: the natural language toolkit.
\newblock In \emph{Proceedings of the COLING/ACL on Interactive presentation
  sessions}, pages 69--72. Association for Computational Linguistics, 2006.

\bibitem[Ioffe and Szegedy(2015)]{ioffe2015batch}
Sergey Ioffe and Christian Szegedy.
\newblock Batch normalization: Accelerating deep network training by reducing
  internal covariate shift.
\newblock In \emph{International Conference on Machine Learning}, pages
  448--456, 2015.

\bibitem[Kim(2014)]{kim2014convolutional}
Yoon Kim.
\newblock Convolutional neural networks for sentence classification.
\newblock \emph{arXiv preprint arXiv:1408.5882}, 2014.

\bibitem[Mikolov et~al.(2013)Mikolov, Sutskever, Chen, Corrado, and
  Dean]{mikolov2013distributed}
Tomas Mikolov, Ilya Sutskever, Kai Chen, Greg~S Corrado, and Jeff Dean.
\newblock Distributed representations of words and phrases and their
  compositionality.
\newblock In \emph{Advances in neural information processing systems}, pages
  3111--3119, 2013.

\bibitem[Roberts et~al.(2014)Roberts, Kilicoglu, Fiszman, and
  Demner-Fushman]{roberts2014automatically}
Kirk Roberts, Halil Kilicoglu, Marcelo Fiszman, and Dina Demner-Fushman.
\newblock Automatically classifying question types for consumer health
  questions.
\newblock In \emph{AMIA Annual Symposium Proceedings}, volume 2014, page 1018.
  American Medical Informatics Association, 2014.

\bibitem[Srivastava et~al.(2014)Srivastava, Hinton, Krizhevsky, Sutskever, and
  Salakhutdinov]{srivastava2014dropout}
Nitish Srivastava, Geoffrey~E Hinton, Alex Krizhevsky, Ilya Sutskever, and
  Ruslan Salakhutdinov.
\newblock Dropout: a simple way to prevent neural networks from overfitting.
\newblock \emph{Journal of machine learning research}, 15\penalty0
  (1):\penalty0 1929--1958, 2014.

\bibitem[Wang and Nyberg(2015)]{wang2015cmu}
Di~Wang and Eric Nyberg.
\newblock Cmu oaqa at trec 2015 liveqa: Discovering the right answer with
  clues.
\newblock Technical report, Carnegie Mellon University Pittsburgh United
  States, 2015.

\bibitem[Wang and Nyberg(2016)]{wang2016cmu}
Di~Wang and Eric Nyberg.
\newblock Cmu oaqa at trec 2016 liveqa: An attentional neural encoder-decoder
  approach for answer ranking.
\newblock In \emph{TREC}, 2016.

\end{thebibliography}

\end{document}